\documentclass{mva_style}
\usepackage{graphicx}
\usepackage{epsfig,graphicx}
\usepackage{graphicx}
\usepackage{epsfig}
\usepackage{latexsym}
\usepackage{amsfonts}
\usepackage{here}
\usepackage{rawfonts}
\usepackage[T1]{fontenc}
\usepackage{calc}
\usepackage{url}
\usepackage{enumerate}
\usepackage{color}
\usepackage[tbtags]{amsmath}
\usepackage{amssymb}
\usepackage{upref}
\usepackage{times}
\usepackage{amsmath}
\usepackage{graphics} 
\usepackage{cite}
\usepackage{booktabs} 
\usepackage{IEEEtrantools}
\usepackage{lipsum,subcaption}
\usepackage{array}
\usepackage{xcolor}
\makeatletter
\newcommand*{\@rowstyle}{}
\newcommand*{\rowstyle}[1]{
  \gdef\@rowstyle{#1}%
  \@rowstyle\ignorespaces%
}
\newcolumntype{=}{
  >{\gdef\@rowstyle{}}%
}
\newcolumntype{+}{
  >{\@rowstyle}%
}
\makeatother

\newcommand{\be}{\begin{enumerate}}
\newcommand{\ee}{\end{enumerate}}
\newcommand{\beqarr}{\begin{eqnarray}}
\newcommand{\beqarrn}{\begin{eqnarray*}}
\newcommand{\eeq}{\end{equation}}
\newcommand{\eeqarr}{\end{eqnarray}}
\newcommand{\eeqarrn}{\end{eqnarray*}}
\newcommand{\bflr}{\begin{flushright}\vspace{-0.2in}}
\newcommand{\eflr}{\end{flushright}}
\newcommand{\bsub}{\begin{subequations}}
\newcommand{\esub}{\end{subequations}}
\newcommand{\barr}{\begin{array}}
\newcommand{\earr}{\end{array}}

\finalcopy 
\begin{document}
\bstctlcite{IEEEexample:BSTcontrol}
\title{Performance Evalution of 3D Keypoint Detectors and Descriptors for Plants Health Classification}
\author{
 Shiva Azimi\\
 Department of Electrical Engineering\\
 Indian Institute of Technology-Delhi\\
 Hauz Khas, New Delhi 110016, India\\
 {\tt shiva.azimi@yahoo.com}\\
 \and
 Brejesh Lall\\
 Department of Electrical Engineering\\
 Indian Institute of Technology-Delhi\\
Hauz Khas, New Delhi 110016, India\\
 {\tt brejesh@ee.iitd.ac.in}\\
 \and
 Tapan K. Gandhi\\
 Department of Electrical Engineering\\
 Indian Institute of Technology-Delhi\\
 Hauz Khas, New Delhi 110016, India\\
 {\tt tgandhi@ee.iitd.ac.in }\\
}
\maketitle
\section*{\centering Abstract} \vspace{-0.3cm}
\textit{Plant Phenomics based on imaging based techniques can be used to 
monitor the health and the diseases of plants and crops. The use of 3D data for plant phenomics is a recent phenomenon. However, since 3D point cloud contains more information than plant images, in this paper, we
compare the performance of different keypoint detectors and local feature descriptors combinations for the plant growth stage and it's growth condition classification based on 3D point clouds of the plants. We have also implemented a modified form of 3D SIFT descriptor, that is invariant to rotation and is computationally less intense than most of the 3D SIFT descriptors reported in the existing literature. The performance is evaluated in terms of the classification accuracy and the results are presented in terms of accuracy tables. We find the ISS-SHOT and the SIFT-SIFT combinations consistently perform better and Fisher Vector (FV) is a better encoder than Vector of Linearly Aggregated (VLAD) for such applications. It can serve as a better modality.
}
\section{INTRODUCTION}  \vspace{-0.3cm}
Increasing world population and the loss of arable land due to various climatic and man made factors have necessitated the development of modern tools and techniques for increasing crop yield \cite{c1}. 
Plant phenotyping is a field of plant science dealing with the measurement of phenomes and how they change in reaction to genetic and environmental changes.\\
\indent In recent years, non-invasive image-based high-throughput plant phenotyping has emerged as an important field of computer vision research \cite{c6}. Measurements of plant features such as biomass, size and inclination, and leaf width, length, and area can be done using two dimensional (2D) \cite{panwar2014imaging} or three dimensional (3D) models of the plant at different scales \cite{srivastava2017drought}.
However, approaches based on 2D models have many limitations since features such as the angle, thickness, and orientation could not be represented using such models. To overcome these issues associated with 2D models, 3D models could be used for plant phenotyping. \\
\indent 3D modeling for plant phenotyping consists of two steps \cite{c6}. The first step consists of localization and mapping. Here,
we define the sensors' poses and compute the scene 3D model. 
The second step is the understanding step. Here, we detect the plant, segment and classify its parts and perform the measurements. To do this, we need to extract features of interest in the plant. The features can be global or local, however local features have been proven more successful in vision tasks like 3D object categorization and recognition.\\
\indent Extraction of local features has two main steps of detection of keypoints and description of patches around these keypoints.
In the keypoints detection stage, points having rich information content identified. Detection of keypoints (or interest points) is important for reducing the amount of computation required in computer vision applications dealing with 3D point clouds because of huge amount of data points present in such point clouds and the high computational cost of descriptors. Once the keypoints are detected, descriptors are extracted for these keypoints.
A number of 3D keypoints detectors and descriptors such as Harris \cite{c8}, Scale-Invariant feature transform (SIFT) \cite{c9}, intrinsic shape signatures (ISS) \cite{c10}, signatures of histograms (SHOT) \cite{c12}, etc., have been proposed in the literature.\\
\indent This work is motivated by the need to quantitatively compare the performance of different keypoint detectors and local feature descriptors for plant health and
growth monitoring. Although, there are studies in the existing literature comparing the performance of 3D keypoint detectors and/or feature descriptors, they all almost either compare either the keypoints detectors or the feature descriptors in certain specific applications \cite{srivastava2019deeppoint3d}. To the best of our knowledge, there is no work in the existing literature that deals specifically with the performance evaluation of the keypoints detectors and the local feature descriptors as applied to 3D point cloud data in the context of plant health classification and plant growth analysis. Working with 3D point cloud data in the context of plant health classification and plant stress analysis offers special challenges: \vspace{-0.25cm}
\begin{itemize}
\item
Plants are made up of very fine structures that makes it very difficult to make a perfect scan. Thus the keypoint detectors 
local feature descriptors have to be robust to noise and holes present in the point cloud data.
\item
The lighting conditions in farming keep changing, offering different illumination conditions, thus, making it very difficult to use the color information for decision making.
\end{itemize}
\vspace{-0.15cm}
In summary, in this paper, we consider three most common 3D keypoints detectors, viz, Harris3D, 3D SIFT and ISS
along with two prominent local feature descriptors, viz, 3D SIFT and SHOT. We have also modified the 3D SIFT descriptor, such that it is invariant to rotation. We provide a quantitative comparison and analysis of these and investigate how they perform in combinations.\\
\indent The remainder of the paper is organized as follows. Section II
gives a brief overview of the related work. We give a description of the keypoints detectors and the local feature descriptors that we have considered for evaluation in Section III. performance evaluation is provided in Section IV. Finally, in Section V, we provide some perspectives and conclude the paper.
\section{Related Works}
\vspace{-0.3cm}
A number of methods have been proposed for creating 3D models for plants. Early approaches for 3D digitization of plants used laser scanning using LiDAR devices \cite{c16}, Time-of-Flight cameras \cite{c17}, Multi-View Stereo (MVS), etc. 
A 3D reconstruction of plant using MVS and free moving cameras was proposed in \cite{c6} for solving the occlusion problem. \\
\indent After making suitable 3D reconstruction of plants with good resolution, we can extract features of interest in the plants. Different traits like plant volume, leaf area, and stem length can be estimated  by using 3D plant model. Wei et al. obtained some volumetric features after making 3D reconstruction of plant by silhouette method \cite{c25}. Extracting features from surfaces meshes is developed recently. Segmented mesh method for computing plant information such as stem length, leaf length and width was proposed in \cite{C24}. \\
\indent However, there are not enough work in the existing literature about descriptors and detectors for plant phenotyping. In addition, it is not clear which of them suitable for which work and application. Most of the work dealing with the performance of the keypoints detectors and local feature descriptors, either evaluate the keypoints detectors or the descriptors, in a particular context. Furthermore, it is not clear how different keypoints detectors and descriptors work in combinations.
F.Tombari evaluated 3D keypoints detectors for object recognition in case of occlusion and clutter \cite{c26}.  
In \cite{c7}, the authors presented a comparison between 3D point detectors and descriptors in the context of point cloud fusion (point cloud registration or alignment). Y. Guo et al. \cite{c34} presented a comprehensive evaluation of 3D local feature descriptors on eight different kind of datasets. \vspace{-0.3cm}
\section{3D Keypoints Detectors and Descriptors} \vspace{-0.3cm}
In this section, we briefly describe the 3D keypoints detectors and the 3D local feature descriptors that we have considered for performance evaluation in this paper. \vspace{-0.3cm}
\subsection{3D Keypoints Detectors} \vspace{-0.24cm}
\subsubsection{Harris3D}   \vspace{-0.3cm}
The 3D-Harris detector is the extension of the 2D corner
detection method of Harris and Stephens \cite{c8}, and works
by taking the first order derivatives along two orthogonal directions on
the 3D surface. The
derivatives are obtained by fitting a quadratic surface to the
neighborhood of a vertex. Unlike the 2D detector, the image gradients in the covariance matrix are replaced by surface normals. To find the keypoints, a Hessian matrix of the intensity is used around each point. This matrix is then smoothed by an isotropic Gaussian filter. 
\subsubsection{ISS} \vspace{-0.3cm}
ISS keypoint detector \cite{c10} relies on region-wise quality measurements for 3D  object recognition which is designed to be stable, repeatable, informative, and discriminative, ensuring highly accurate 3D shape matching and recognition. This method uses the magnitude of the smallest eigenvalue and the ratio between two successive eigenvalues.
\subsubsection{SIFT3D}  \vspace{-0.3cm}
The SIFT keypoint detector finds local extrema in a Difference-of-Gaussians (DoG) scale-space \cite{c9}. 3D SIFT is an extension of the 2D SIFT to 3D. First a density function is approximated by sampling the data regularly in space. A scale space is then built over the density function, and a search is done for local maxima of the Hessian determinant. To generate scale space the input cloud is convolved with a number of Gaussian filters whose standard deviations differ by a fixed scale factor. 
The 3D SIFT keypoints are positioned at the scale-space extrema of the DoG function. All the keypoints with low curvature values are rejected to get stable results.
 \begin{figure*}[ht!]
    \centering
    \begin{subfigure}[b]{0.25\textwidth}
        \includegraphics[height=1.4cm, width=3cm]{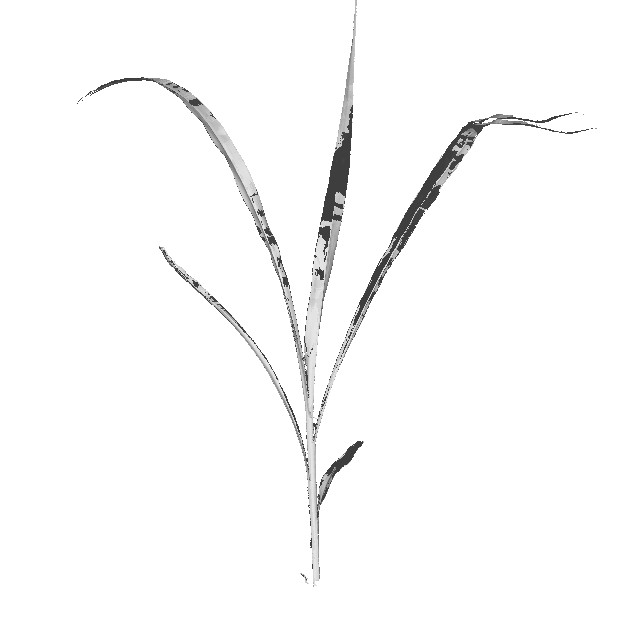}
        \caption{Sorghum (control)}
        \label{fig:gull}
    \end{subfigure}
    \begin{subfigure}[b]{0.25\textwidth}
        \includegraphics[height=1.4cm, width=3cm]{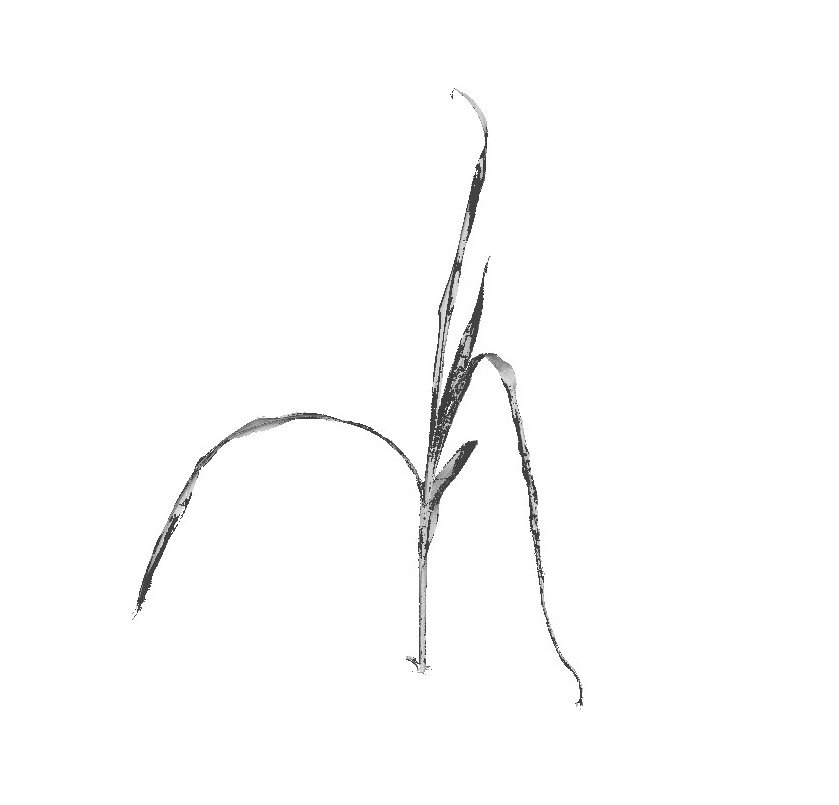}
        \caption{Sorghum (heat)}
        \label{fig:tiger}
    \end{subfigure}
    \begin{subfigure}[b]{0.25\textwidth}
        \includegraphics[height=1.4cm, width=3cm]{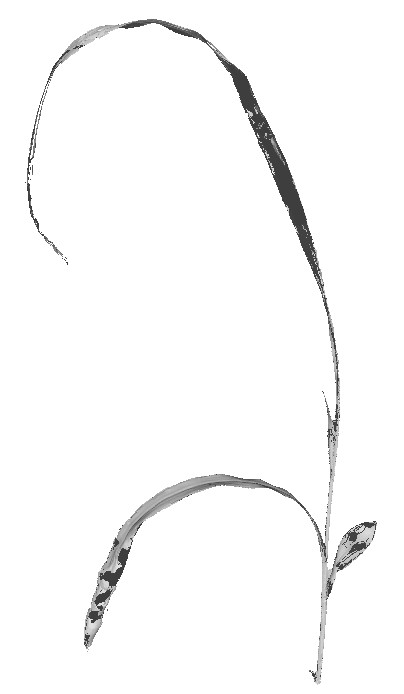}
        \caption{Sorghum (shade)}
        \label{fig:mouse}
    \end{subfigure} \vspace{-0.3cm}
    \bigskip
      \begin{subfigure}[b]{0.25\textwidth}
        \includegraphics[height=1.4cm, width=3cm]{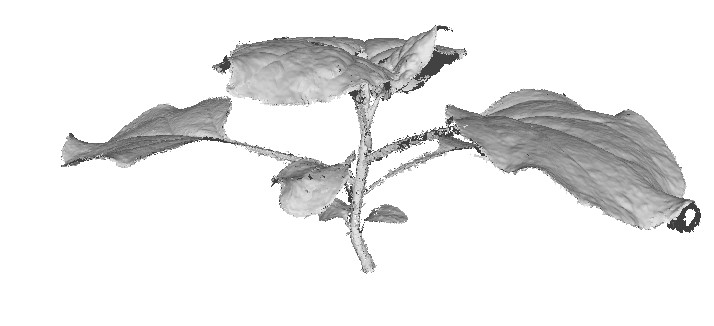}
        \caption{Tobacco (control)}
        \label{fig:gull}
    \end{subfigure}
    \begin{subfigure}[b]{0.25\textwidth}
        \includegraphics[height=1.4cm, width=3cm]{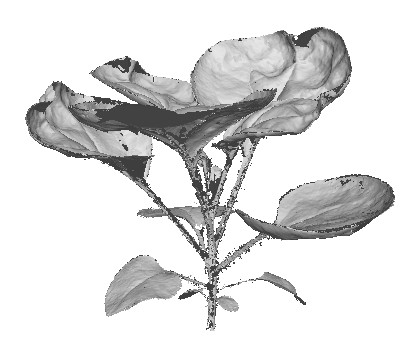}
        \caption{Tobacco (heat)}
        \label{fig:tiger}
    \end{subfigure}
    \begin{subfigure}[b]{0.25\textwidth}
        \includegraphics[height=1.4cm, width=3cm]{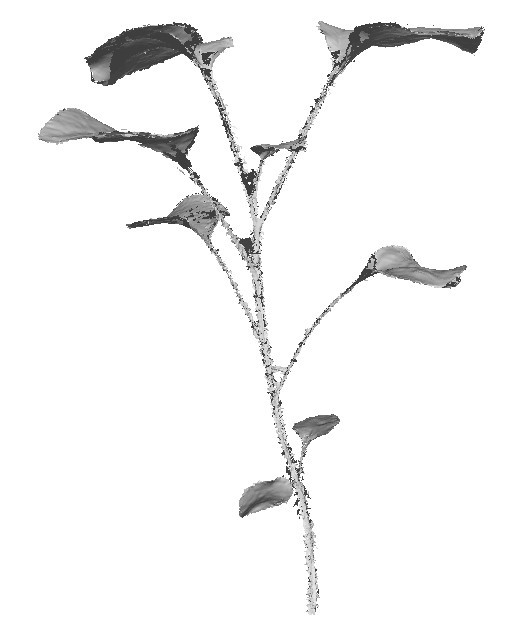}
        \caption{Tobacco (shade)}
        \label{fig:mouse}
    \end{subfigure} \vspace{-0.3cm}
      \bigskip
      \begin{subfigure}[b]{0.25\textwidth}
        \includegraphics[height=1.4cm, width=3cm]{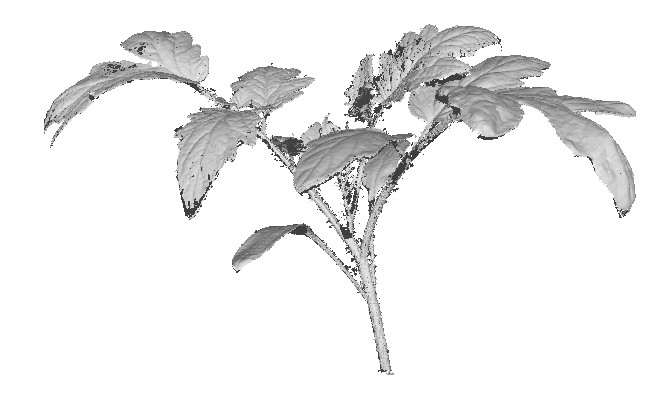}
        \caption{Tomato (control)}
        \label{fig:gull}
    \end{subfigure}
    \begin{subfigure}[b]{0.25\textwidth}
        \includegraphics[height=1.4cm, width=3cm]{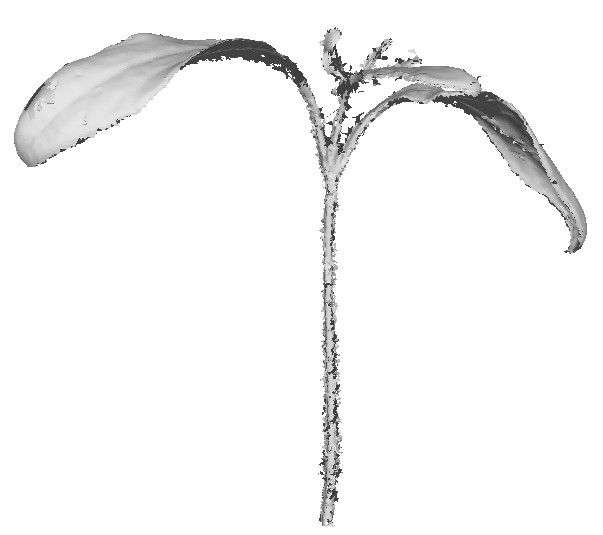}
        \caption{Tomato (heat)}
        \label{fig:tiger}
    \end{subfigure}
    \begin{subfigure}[b]{0.25\textwidth}
        \includegraphics[height=1.4cm, width=3cm]{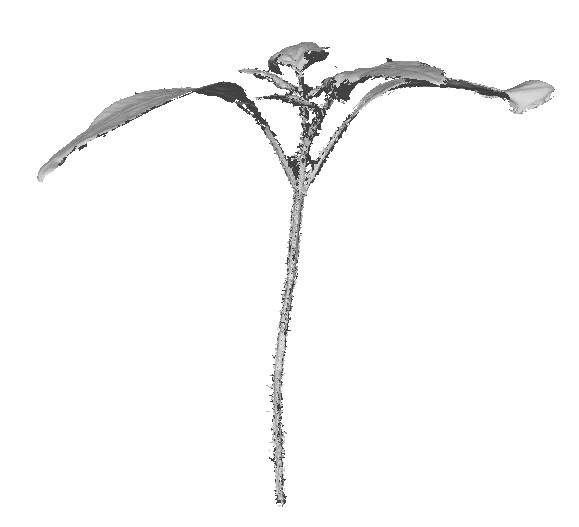}
        \caption{Tomato (shade)}
        \label{fig:mouse}
    \end{subfigure}
    \vspace{-.4cm}
    \caption{Sample point clouds from the dataset for the three conditions: control, heat and shade}\label{fig:animals} \vspace{-0.3cm}
\end{figure*}  
\vspace{-.15cm}
\subsection{3D Local Feature Descriptors} \vspace{-0.2cm}
\subsubsection{SIFT3D} \vspace{-0.3cm}
3D SIFT involves the same functional steps as for 2D SIFT. In our case, we have slightly modified the steps to make the descriptors invariant to rotation and to speed up the description process. For example, we take the entire $r-point$ neighborhood around each keypoint rather than dividing it into $n \times n \times n$ subregions (with $n=4$, as in Lowe's 2D SIFT). Using these $r$-point a 2D-histogram is produced by grouping the angles $\theta$ and $\phi$ into 10 degree angular bins. 
Similar to 2D SIFT, a regional Gaussian weighting of ${e^{{{ - 2d} \mathord{\left/
 {\vphantom {{ - 2d} {{R_{\max }}}}} \right.
 \kern-\nulldelimiterspace} {{R_{\max }}}}}}$ for the points that are at a distance
$d$ is applied to the histogram. Here $R_{max}$ represents the distance of the neighborhood point farthest from the center.
The prominent azimuth $\alpha$ and elevation $\beta$ for a keypoint are given by the peaks of the 2D-histogram. 
Each keypoint $p$ is described by its location ${\bf{p}} \buildrel \Delta \over = {\left[ {{x_p},{y_p},{z_p}} \right]^{\,t}}$, scale $\sigma_p$, and orientation angles $\alpha_p$ and $\beta_p$. To ensure rotation invariance of the descriptor, the $r$-points $p_i$, $i \in 1, 2, {\cdots}, r$, with coordinates ${\bf{p}_i} \buildrel \Delta \over = {\left[ {{x_i},{y_i},{z_i}} \right]^{\,t}}$ around the keypoint of interest $p$ are at first rotated in the dominant orientation of $p$ using the transformation
\begin{eqnarray}
{\bf{p}}_{\bf{i}}^{\bf{'}} = \left[ {\begin{array}{*{20}{c}}
{\cos {\alpha _p}\cos {\beta _p}}&{ - \sin {\alpha _p}}&{ - \cos {\alpha _p}\sin {\beta _p}}\\
{\sin {\alpha _p}\cos {\beta _p}}&{\cos {\alpha _p}}&{ - \sin {\alpha _p}\sin {\beta _p}}\\
{\sin {\beta _p}}&0&{\cos {\beta _p}}
\end{array}} \right] \cdot {{\bf{p}}_{\bf{i}}}.
\label{eq1}
\end{eqnarray}
Then, the normal vector to the neighborhood $\bf{n}$ is calculated at the current key point $\bf{p}$. For each of these rotated points $\bf{p_i}^{\bf{'}}$ in the $r$-points neighborhood of $\bf{p}$, the vector ${{\bf{p}}^{\bf{'}}}{\bf{p}}_{\bf{i}}^{\bf{'}}$ is computed, where ${{\bf{p}}^{\bf{'}}}$ is the keypoint $\bf{p}$ rotated according to the transformation (\ref{eq1}). Then, we calculate the angle $\delta$ as
\begin{IEEEeqnarray}{c}
\delta  = {\cos ^{ - 1}}\left( {\frac{{{{\bf{p}}^{\bf{'}}}{\bf{p}}_{\bf{i}}^{\bf{'}} \cdot {\bf{n}}}}{{\left| {{{\bf{p}}^{\bf{'}}}{\bf{p}}_{\bf{i}}^{\bf{'}}} \right| \cdot \left| {\bf{n}} \right|}}} \right).
\label{eq2}
\end{IEEEeqnarray}
Thus, every keypoint along with its neighborhood is represented by a 4-tuple $\left(m, \theta, \phi, \delta \right)$. The $\left(\theta, \phi, \delta \right)$ space is then  divided into $45$ deg bins, and these bins are populated by adding up the corresponding values with a Gaussian weighting of ${e^{{{ - 2d} \mathord{\left/
 {\vphantom {{ - 2d} {{R_{\max }}}}} \right.
 \kern-\nulldelimiterspace} {{R_{\max }}}}}}$. 
The azimuth angle $\theta \in \left[0, 360\right]$ deg is divided into 8 bins of 45 deg; the elevation angle $\phi \in \left[{{- 90}, 90}\right]$
deg is divided into 4 bins of 45 deg; and $\delta \in \left[0, 180\right]$ deg which is also split into 4 bins of 45 deg. Thus, the size of the 3D SIFT descriptors in our case is $4 \times 4 \times 8=128$. Each 3D SIFT descriptor is normalized to unity.
\vspace{-0.2cm}
\subsubsection{SHOT} \vspace{-0.25cm}
Motivated by the SIFT descriptor, Signatures
of Histograms of Orientations (SHOT), was proposed to take the advantages of
both signature as well as histogram based methods simultaneously \cite{c12}. In SHOT, a unique and repeatable local reference frame is computed for each keypoint using an eigenvalue decomposition of the modified neighborhood covariance matrix $\bf{C}$.
The sample points
$\bf{q_i}$ that lie in the support region of radius $\bf{r}$ are weighted based on their
distances from the point $\bf{q}$ under consideration, as shown below
\begin{IEEEeqnarray}{c}
{\bf{C}} = \frac{1}{{\sum\nolimits_{i:{d_i} \le {\bf{r}}} {\left( {{\bf{r}} - {d_i}} \right)} }}\sum\limits_{i:{d_i} \le {\bf{r}}} {\left( {{\bf{r}} - {d_i}} \right)} \left( {{{\bf{q}}_{\bf{i}}}{\bf{ - q}}} \right){\left( {{{\bf{q}}_{\bf{i}}}{\bf{ - q}}} \right)^t},\IEEEeqnarraynumspace
\label{eq3}
\end{IEEEeqnarray}
where, ${d_i} = {\left\| {{q_i} - q} \right\|_2}$. 
Using the reference frame, a spherical grid is built around the input point. This grid divides supporting points into grid cells. At each grid sector, a weighted cosine of the relative normal angle is calculated, and the result is binned into a local histogram for that cell.
SHOT combines all local histograms into one descriptor of length 352. In the last stage, the descriptor is normalized to unity.
\begin{figure*}[ht!]
      \centering
     \includegraphics[height=3.9cm, width=11.8cm]{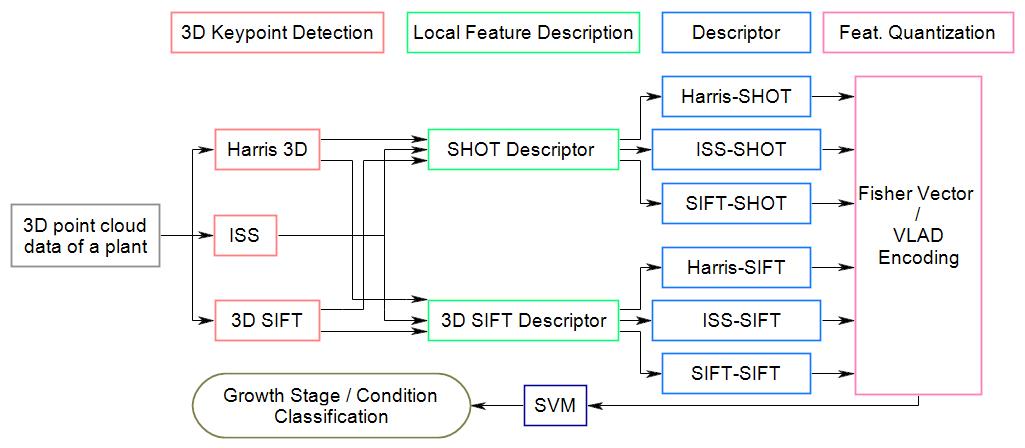}
     \vspace{-.2cm}
      \caption{Representation of the performance evaluation of 3D keypoint detectors and descriptors}  
      \label{figurelabel} \vspace{-0.3cm}
   \end{figure*}  
\subsection{Fisher Vector (FV) and Vector of Linearly Aggregated Descriptors (VLAD) Encoding} \vspace{-0.3cm}
Once we have extracted 3D features from point cloud of the plants, the next step is to classify them into different categories. Since, the number of keypoints for different point cloud are different, we need to quantize the 3D descriptors. In this paper, we have experimented with both FV \cite{c47} and VLAD \cite{c33} for encoding the 3D features. \\
\indent A FV is a statistics representing the distribution of a set of vectors, in our case, a set of local surface descriptors. The FV encoding employs Gaussian Mixture Model (GMM) to the local descriptors to construct a dictionary of $k$ words. Once the dictionary is created, the FV encodes the gradients of the log-likelihood of the features under the GMM, with respect to the GMM parameters and compare $N$ descriptors to $k$ visual words. \\
\indent VLAD is another feature encoding and pooling method used to encode a set of local feature descriptors using a dictionary built using k-means clustering, unlike the FV, that used GMM. \\
\indent The FV and VLAD representations offer many advantages compared with other methods. They can be computed from smaller vocabularies, and therefore require less computation. Furthermore, FV and VLAD perform well even with simple linear classifiers as they don't come up with curvy and complex boundaries between classes. This is a significant advantage as linear classifiers are very efficient to learn and easy to evaluate.  \\
\indent In this work, we directly quantize the features using FV and VLAD encoding. Since the size of the generated codebook is small, we use Support Vector Machine (SVM), a linear classifier, for training and classification. 
\subsection{SVM} \vspace{-0.3cm}
A SVM is a supervised learning model, formally defined by a separating hyperplane.
Given the set of labeled training data (supervised learning), the algorithm creates an optimal hyperplane that categorizes new test data, making it a non-probabilistic binary linear classifier. There are two common approaches for extending SVM from a binary classifier to a multi-class classifier: one-versus-one and one-vs-all. In this paper, we have used the one versus all approach. 
\section{Experimental Evaluation And Discussion} \vspace{-0.3cm}
In this section we provide the quantitative as well as qualitative performance of different keypoint detector and descriptor combinations in context of plant health classification.
\subsection{Dataset Used}
\vspace{-.25cm}
In this work we used the dataset containing 3D scans of plant shoot architectures provided by Salk Institute of Biological Science \cite{c49}, as there are not many publicly available  dataset of 3D point clouds of plants prepared to study the growth and health of plants under different environmental conditions. This dataset contains $559$ 3D plant shoot architectures from 4 species (Arabidopsis, tomato, tobacco, and sorghum) scanned under multiple conditions (ambient light (control), high-heat, high-light, shade, drought). Each plant was scanned every 1 or 2 days through roughly 20 days of development.\\
\indent In our work we have used only sorghum, tobacco and tomato plants in three different  conditions of control, shade and heat because this data has the same categories and the same number of replicates which makes it easier to compare the results.\vspace{-0.1cm}
\subsection{Experiments and Results}  \vspace{-0.3cm}
All our experiments were performed using MATLAB and/or the point cloud library (PCL) (See Table 1). Each plant had three replicates in each category, viz, control, heat and shade. All the point clouds across all the replicates for each plant were divided in to categories based on the growth stage, and also on the growth condition. For classification based on the growth condition, the point clouds for each plant were grouped in to three categories: control, heat and shade across all the replicates. For growth stage, we grouped the point clouds for each plant in to three categories: stage 1, stage 2, and stage 3. For this, the point clouds across all the growth conditions were grouped into the three stages, with stage three representing the most mature plants. For each point cloud in the dataset, we extract the keypoints using three most common 3D keypoint detectors, viz, Harris 3D, ISS and 3D SIFT, giving us three sets of keypoints. We then use 
SHOT and 3D SIFT descriptors to describe the local surface around these sets of keypoints, giving us a total of six detector-descriptor combinations for each point cloud. This is followed by the quantization of these descriptors using FV and VLAD encoding. In this work,
We have performed the evaluation for the FV and VLAD encoded descriptors separately.
Once the descriptors have been extracted and quantized, SVMs are trained to learn plant categories and to perform the plant classification for both growth stage and growth condition. The involved steps are shown in Fig. 1.
For the classification step, we use $80\%$ of the feature vectors for training the SVM. Linear
SVM classifier with one-versus-all approach is applied for its efficiency and simplicity.

The performance evaluation of a detector - descriptor pair is done in terms 
of classification accuracy. The accuracy is calculated as
the percentage of number of correct classifications to the total number of test inputs for respective categories.
\begin{table}[ht!]
\caption{ Implementation Platforms of the Used Techniques} \vspace{-0.5cm}
\label{tableI}
\begin{center}
\begin{tabular}{c | c}
\hline
\hline
\makebox[10mm]{Technique} & \makebox[10mm]{Platform}\\
\hline
Harris3D Detector & PCL\\
\hline
ISS Detector & PCL\\%
\hline
SIFT Detector & PCL\\%
\hline
3D SIFT Descriptor & MATLAB\\%
\hline
SHOT Descriptor & PCL\\%
\hline
VLAD Encoder & MATLAB\\%
\hline
Fisher Vector & MATLAB\\%
\hline
\hline
\end{tabular}
\end{center}
\end{table}
\begin{table}[ht!]
\caption{Classification Accuracy ($\%$) for Sorghum (both growth stage and growth conditions)} \vspace{-0.15cm}
\centering 
\begin{tabular}{c | c c c c} 
\hline 
\hline
\makebox[10mm]{Pair} &\multicolumn{2}{c}{\makebox[10mm]{Accuracy (FV)}} &\multicolumn{2}{c}{\makebox[10mm]{Accuracy (VLAD)}}\\ 
\hline 
 & \makebox[12mm]{Conditon} & \makebox[10mm]{Stage} & \makebox[10mm]{Condition} & \makebox[10mm]{Stage}\\ 
\hline 
Harris-SHOT & 62.96 &  \rowstyle{\color{red}}59.26 &  \rowstyle{\color{red}}66.67 & 66.67\\ 
ISS-SHOT & 85.18 & \rowstyle{\color{blue}}77.78 & 85.18 & \rowstyle{\color{blue}}81.48\\
SIFT-SHOT & 74.07 & 74.07 & 70.37 & 70.37\\
Harris-SIFT &  \rowstyle{\color{red}}59.26 & 59.26 & 74.07 &  \rowstyle{\color{red}}59.26\\
ISS-SIFT & 81.48 & 77.78 & 85.18 & 70.37\\
SIFT-SIFT & \rowstyle{\color{blue}}85.18 & 62.96 & \rowstyle{\color{blue}}88.89 & 62.96\\ 
\hline 
\hline
\end{tabular}
\label{table2}
\end{table}
\vspace{-0.05cm}
\begin{table}[ht!]
\caption{Classification Accuracy ($\%$) for Tobacco (both growth stage and growth conditions)} \vspace{-0.15cm}
\centering 
\begin{tabular}{c | c c c c} 
\hline 
\hline
\makebox[10mm]{Pair} &\multicolumn{2}{c}{\makebox[10mm]{Accuracy (FV)}} &\multicolumn{2}{c}{\makebox[10mm]{Accuracy (VLAD)}}\\ 
\hline 
 & \makebox[12mm]{Conditon} & \makebox[10mm]{Stage} & \makebox[10mm]{Condition} & \makebox[10mm]{Stage}\\
\hline 
Harris-SHOT &
\rowstyle{\color{red}}
 55.56 & \rowstyle{\color{red}} 59.26 &  \rowstyle{\color{red}}62.96& \rowstyle{\color{red}}55.56\\ 
ISS-SHOT & 81.48 & \rowstyle{\color{blue}}85.18 & 81.48 & \rowstyle{\color{blue}}77.78\\
SIFT-SHOT & 81.48 & 71.78 & \rowstyle{\color{blue}}85.18 & 77.78\\
Harris-SIFT & 66.67 & 62.96 & 70.37 & 62.96\\
ISS-SIFT & 81.48 & 70.37 & 74.07 & 59.26\\
SIFT-SIFT &  \rowstyle{\color{blue}}
 85.18 & 62.96 & 77.78 & 62.96\\[1ex] 
\hline 
\hline
\end{tabular}
\label{table3}
\end{table}
\begin{table}[ht!]
\caption{Classification Accuracy ($\%$) for Tomato (both growth stage and growth conditions)} \vspace{-0.15cm}
\centering 
\begin{tabular}{c | c c c c} 
\hline 
\hline
\makebox[10mm]{Pair} &\multicolumn{2}{c}{\makebox[10mm]{Accuracy (FV)}} &\multicolumn{2}{c}{\makebox[10mm]{Accuracy (VLAD)}}\\ 
\hline 
 & \makebox[12mm]{Conditon} & \makebox[10mm]{Stage} & \makebox[10mm]{Condition} & \makebox[10mm]{Stage}\\
\hline 
Harris-SHOT & \rowstyle{\color{red}}70.37 & \rowstyle{\color{red}}66.67 & 74.07 & \rowstyle{\color{red}}66.67\\ 
ISS-SHOT & 92.60 & 77.78 & 85.18 & \rowstyle{\color{blue}}81.48\\
SIFT-SHOT & 88.89 &  \rowstyle{\color{blue}}81.48 &  \rowstyle{\color{blue}}88.89 & 77.78\\
Harris-SIFT & 81.48 & 74.07 & \rowstyle{\color{red}}70.37 & 66.67\\
ISS-SIFT & 96.30 & 70.37 & 88.89 & 70.37\\
SIFT-SIFT &  \rowstyle{\color{blue}}96.30 & 74.07 & 85.18 & 77.78\\[1ex] 
\hline 
\hline
\end{tabular}
\label{table4}
\end{table}
\begin{table}[ht!]
\caption{Classification Accuracy ($\%$) Averaged Over All Three Plants (both growth stage and growth conditions)} \vspace{-0.15cm}
\centering 
\begin{tabular}{c | c c c c} 
\hline 
\hline
\makebox[10mm]{Pair} &\multicolumn{2}{c}{\makebox[10mm]{Accuracy (FV)}} &\multicolumn{2}{c}{\makebox[10mm]{Accuracy (VLAD)}}\\ 
\hline 
 & \makebox[12mm]{Conditon} & \makebox[10mm]{Stage} & \makebox[10mm]{Condition} & \makebox[10mm]{Stage}\\
\hline 
Harris-SHOT & \rowstyle{\color{red}}62.96 & \rowstyle{\color{red}}61.723 & \rowstyle{\color{red}}67.9 & \rowstyle{\color{red}}62.96\\ 
ISS-SHOT & 86.42 &  \rowstyle{\color{blue}}80.25 & 83.95 &  \rowstyle{\color{blue}}80.25\\
SIFT-SHOT & 81.48 & 75.78 & 81.48 & 75.30\\
Harris-SIFT & 69.14 & 65.43 & 71.60 & 62.96\\
ISS-SIFT & 86.42 & 72.84 & 82.71 & 66.67\\
SIFT-SIFT &  \rowstyle{\color{blue}}88.87 & 66.66 &  \rowstyle{\color{blue}}83.95 & 67.90 \\[1ex] 
\hline 
\hline
\end{tabular}
\label{table5}
\end{table} Tables 2, 3 and 4 show the classification accuracies of the six detector-descriptor (in the same order) pairs with FV as well as VLAD encoding for both condition-wise and growth-wise classification for 
sorghum, tobacco and tomato, respectively. The detector-descriptor pair giving the worst and the best results in each column is highlighted in red and blue, respectively. One important observation regarding the physical structure of the plants is that they vary significantly in the shape of the leaves and their curvature variations. Sorghum has thin leaves and high curvature variations, tobacco has broad leaves and low curvature variations, and tomato, also, has broad leaves (smaller than tobacco) but has a relatively higher curvature variations compared to tobacco. These differences in the shape affect the detection and the description processes, and in turn affect the classification accuracies. Table 2 shows that for sorghum, the Harris keypoints give poor results with both SHOT and SIFT descriptors, while the ISS keypoints perform well with both the descriptors. The SIFT descriptors perform very well with the SIFT keypoints in case of FV as well as VLAD encoding. Similar trends are observed for tobacco and tomato, as seen from Tables 3 and 4, with both the descriptors performing well with the ISS keypoints while performing poorly with Harris keypoints.\\
\indent Table 5 shows the average classification accuracy across all the three plants. Here ISS-SHOT and SIFT-SIFT outperform all other combinations. Also, one can note that the classification accuracies are on an average better with FV encoding as compared to VLAD encoding. One trend that is common across all the tables is that the classification accuracies for growth stage classification are lower than that for growth condition classification. This trend can be attributed to the fact that the similarity between plants growing in the same environmental condition observed over successive days will be usually higher than the similarity between plants that are growing under different environmental conditions.
\section{CONCLUSIONS} \vspace{-0.3cm}
In this paper, we
compared the performance of different keypoint detector-descriptors combinations in context of plant growth classification and it's growth condition using 3D point clouds of the plants. The performance was evaluated in terms of the classification accuracy and the results were presented in terms of accuracy tables. Experimental results show that the ISS keypoints perform much better than the Harris keypoints with both SHOT and SIFT descriptors. Furthermore, the ISS-SHOT combination gives better results in the high curvature cases while the SIFT-SIFT combination performs well in all the cases generally. Also, in general, FV performs better than VLAD encoding. In future works, we intend to prepare our own dataset of 3D point clouds of crops such as rice and wheat and will try to develop and analyze 
novel methods and detector-descriptor combinations tailored for plant phenomics application. We will also explore the use of neural networks and other machine learning methods for applications related to plant phenomics. 
\balance
\bibliographystyle{IEEEtran}
\bibliography{egbib}

\begin{thebibliography}{10}
\providecommand{\url}[1]{#1}
\csname url@samestyle\endcsname
\providecommand{\newblock}{\relax}
\providecommand{\bibinfo}[2]{#2}
\providecommand{\BIBentrySTDinterwordspacing}{\spaceskip=0pt\relax}
\providecommand{\BIBentryALTinterwordstretchfactor}{4}
\providecommand{\BIBentryALTinterwordspacing}{\spaceskip=\fontdimen2\font plus
\BIBentryALTinterwordstretchfactor\fontdimen3\font minus
  \fontdimen4\font\relax}
\providecommand{\BIBforeignlanguage}[2]{{%
\expandafter\ifx\csname l@#1\endcsname\relax
\typeout{** WARNING: IEEEtran.bst: No hyphenation pattern has been}%
\typeout{** loaded for the language `#1'. Using the pattern for}%
\typeout{** the default language instead.}%
\else
\language=\csname l@#1\endcsname
\fi
#2}}
\providecommand{\BIBdecl}{\relax}
\BIBdecl

\bibitem{c1}
A.~J. Challinor, J.~Watson, D.~Lobell, S.~Howden, D.~Smith, and N.~Chhetri, ``A
  meta-analysis of crop yield under climate change and adaptation,''
  \emph{Nature Climate Change}, vol.~4, no.~4, p. 287, 2014.

\bibitem{c6}
T.~T. Santos, L.~V. Koenigkan, J.~G.~A. Barbedo, and G.~C. Rodrigues, ``3d
  plant modeling: localization, mapping and segmentation for plant phenotyping
  using a single hand-held camera,'' in \emph{European Conference on Computer
  Vision}.\hskip 1em plus 0.5em minus 0.4em\relax Springer, 2014, pp. 247--263.

\bibitem{panwar2014imaging}
R.~Panwar, K.~Goyal, N.~Pandey, and N.~Khanna, ``Imaging system for
  classification of local flora of uttarakhand region,'' in \emph{2014
  International Conference on Power, Control and Embedded Systems
  (ICPCES)}.\hskip 1em plus 0.5em minus 0.4em\relax IEEE, 2014, pp. 1--6.

\bibitem{srivastava2017drought}
S.~Srivastava, S.~Bhugra, B.~Lall, and S.~Chaudhury, ``Drought stress
  classification using 3d plant models,'' in \emph{Proceedings of the IEEE
  International Conference on Computer Vision}, 2017, pp. 2046--2054.

\bibitem{c8}
C.~Harris and M.~Stephens, ``A combined corner and edge detector in alvey
  vision conference,'' \emph{Manchester, UK}, 1988.

\bibitem{c9}
D.~G. Lowe, ``Object recognition from local scale-invariant features,'' in
  \emph{Computer vision, 1999. The proceedings of the seventh IEEE
  international conference on}, vol.~2.\hskip 1em plus 0.5em minus 0.4em\relax
  Ieee, 1999, pp. 1150--1157.

\bibitem{c10}
D.-Y. Chen, X.-P. Tian, Y.-T. Shen, and M.~Ouhyoung, ``On visual similarity
  based 3d model retrieval,'' in \emph{Computer graphics forum}, vol.~22,
  no.~3.\hskip 1em plus 0.5em minus 0.4em\relax Wiley Online Library, 2003, pp.
  223--232.

\bibitem{c12}
S.~Salti, F.~Tombari, and L.~Di~Stefano, ``Shot: Unique signatures of
  histograms for surface and texture description,'' \emph{Computer Vision and
  Image Understanding}, vol. 125, pp. 251--264, 2014.

\bibitem{srivastava2019deeppoint3d}
S.~Srivastava and B.~Lall, ``Deeppoint3d: Learning discriminative local
  descriptors using deep metric learning on 3d point clouds,'' \emph{Pattern
  Recognition Letters}, 2019.

\bibitem{c16}
S.~Delagrange and P.~Rochon, ``Reconstruction and analysis of a deciduous
  sapling using digital photographs or terrestrial-lidar technology,''
  \emph{Annals of botany}, vol. 108, no.~6, pp. 991--1000, 2011.

\bibitem{c17}
G.~Aleny{\`a}, B.~Dellen, and C.~Torras, ``3d modelling of leaves from color
  and tof data for robotized plant measuring,'' in \emph{Robotics and
  Automation (ICRA), 2011 IEEE International Conference on}.\hskip 1em plus
  0.5em minus 0.4em\relax IEEE, 2011, pp. 3408--3414.

\bibitem{c25}
W.~Fang, H.~Feng, W.~Yang, L.~Duan, G.~Chen, L.~Xiong, and Q.~Liu,
  ``High-throughput volumetric reconstruction for 3d wheat plant architecture
  studies,'' \emph{Journal of Innovative Optical Health Sciences}, vol.~9,
  no.~05, p. 1650037, 2016.

\bibitem{C24}
A.~Paproki, J.~Fripp, O.~Salvado, X.~Sirault, S.~Berry, and R.~Furbank,
  ``Automated 3d segmentation and analysis of cotton plants,'' in \emph{Digital
  Image Computing Techniques and Applications (DICTA), 2011 International
  Conference on}.\hskip 1em plus 0.5em minus 0.4em\relax IEEE, 2011, pp.
  555--560.

\bibitem{c26}
F.~Tombari, S.~Salti, and L.~Di~Stefano, ``Performance evaluation of 3d
  keypoint detectors,'' \emph{International Journal of Computer Vision}, vol.
  102, no. 1-3, pp. 198--220, 2013.

\bibitem{c7}
R.~H{\"a}nsch, T.~Weber, and O.~Hellwich, ``Comparison of 3d interest point
  detectors and descriptors for point cloud fusion,'' \emph{ISPRS Annals of the
  Photogrammetry, Remote Sensing and Spatial Information Sciences}, vol.~2,
  no.~3, p.~57, 2014.

\bibitem{c34}
Y.~Guo, M.~Bennamoun, F.~Sohel, M.~Lu, J.~Wan, and N.~M. Kwok, ``A
  comprehensive performance evaluation of 3d local feature descriptors,''
  \emph{International Journal of Computer Vision}, vol. 116, no.~1, pp. 66--89,
  2016.

\bibitem{c47}
F.~Perronnin and C.~Dance, ``Fisher kernels on visual vocabularies for image
  categorization,'' in \emph{Computer Vision and Pattern Recognition, 2007.
  CVPR'07. IEEE Conference on}.\hskip 1em plus 0.5em minus 0.4em\relax IEEE,
  2007, pp. 1--8.

\bibitem{c33}
H.~J{\'e}gou, M.~Douze, C.~Schmid, and P.~P{\'e}rez, ``Aggregating local
  descriptors into a compact image representation,'' in \emph{Computer Vision
  and Pattern Recognition (CVPR), 2010 IEEE Conference on}.\hskip 1em plus
  0.5em minus 0.4em\relax IEEE, 2010, pp. 3304--3311.

\bibitem{c49}
\BIBentryALTinterwordspacing
S.~Navlakha, ``3d scans of plant shoot architectures,'' Jul 2017. [Online].
  Available: \url{https://data.mendeley.com/datasets/9k7zctdyhs/1}
\BIBentrySTDinterwordspacing

\end{thebibliography}
\end{document}